%% file: main.tex
\documentclass[letterpaper, 10 pt, conference]{ieeeconf}  % Comment this line out if you need a4paper

\usepackage{graphics} % for pdf, bitmapped graphics files
\usepackage{amsmath} % assumes amsmath package installed
\usepackage{amssymb}  % assumes amsmath package installed

\usepackage{multirow}
\usepackage{booktabs}
\usepackage{makecell}
\usepackage{tabularx}
\usepackage{romannum}
\usepackage{graphicx} % 用于插入图片
\usepackage{caption}  % 用于图片标题设置
\usepackage{subcaption} % 如需插入子图时使用
\usepackage{float} % 用于[H]位置参数
\usepackage{cite}
\usepackage{xcolor}
\usepackage{listings}
\usepackage[pagebackref,breaklinks,colorlinks]{hyperref}

\usepackage{cuted}
\title{\LARGE \bf
Sparse Meets Dense: Correspondence Guided Robotic Manipulation with Rigid-Deformable Interactions
}

\author{
Ziyu Zhu $^{2,1}$ \quad
Yue Chen $^{1}$ \quad
Xirui Liang $^{2,1}$  \quad 
Hojin Bae $^{1}$  \quad 
Yuran Wang $^{1}$ \quad
Zhen Yuan $^{1}$\quad \\
Ruihai Wu $^{\dagger 1}$\quad
Hao Dong $^{\dagger 1}$ \quad  \\
$^1$CFCS, School of Computer Science, PKU \quad 
$^2$School of EECS, PKU\\
$\dagger$Corresponding author
}

\begin{document}

\maketitle

\thispagestyle{empty}
\pagestyle{empty}

%%%%%%%%%%%%%%%%%%%%%%%%%%%%%%%%%%%%%%%%%%%%%%%%%%%%%%%%%%%%%%%%%%%%%%%%%%%%%%%%

\begin{strip}
\vspace{-9mm}
    \centering
    \includegraphics[width=\linewidth]{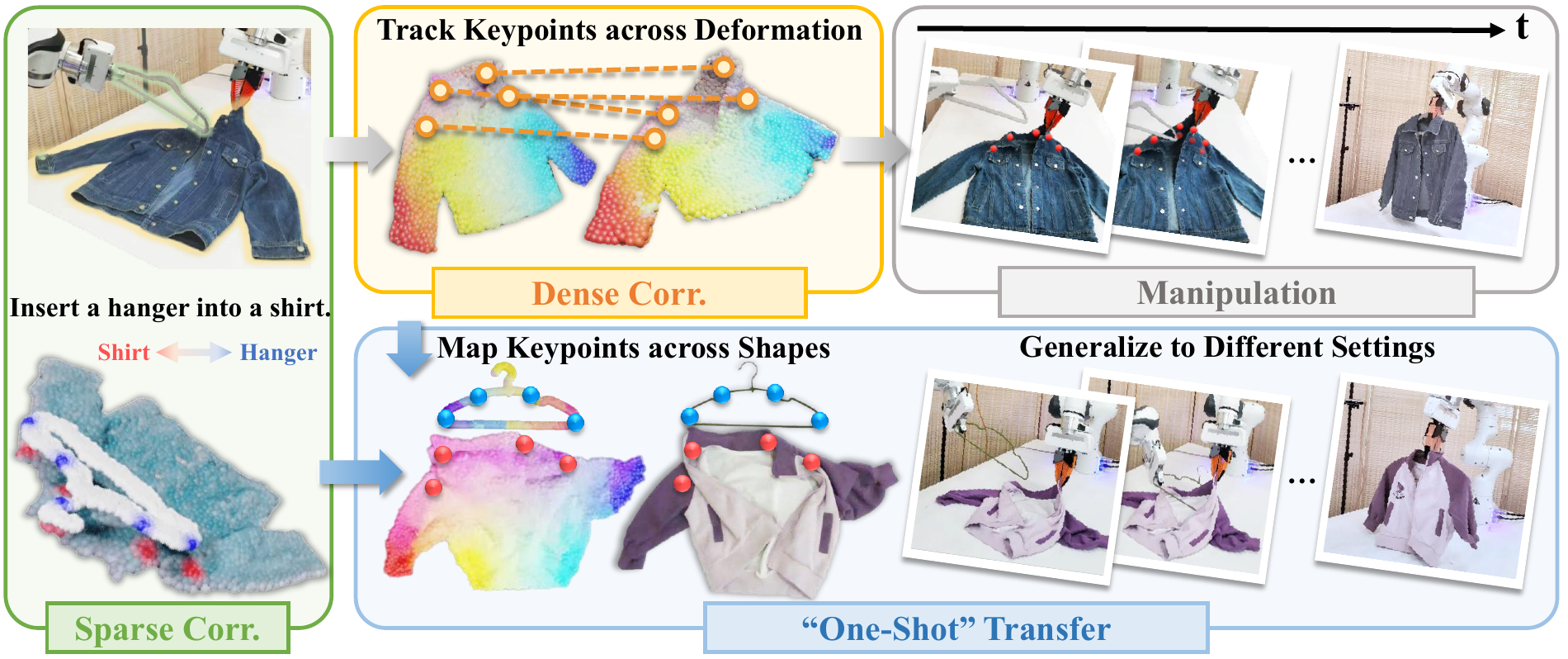}
    \captionof{figure}
    {\textbf{Correspondences for Rigid-Deformable Interactions.} We propose a hybrid correspondence-based representation for rigid-deformable interaction tasks. \textbf{(Left)} Sparse keypoint correspondences capture the interaction information between objects. \textbf{(Top-middle)} For long-horizon manipulation with a sequence of actions, dense correspondences can track keypoint accurately on soft bodies across deformations, and enable one-shot transfer by mapping keypoints to new shapes.}  
    \label{fig:teaser}
% \vspace{-5mm}
\end{strip}

\input{section/0_abstract}

\input{section/1_introduction}
\input{section/2_related_work}

\input{section/3_problem_formulation}
\input{section/4_method}
\input{section/5_experiment}

\input{section/6_conclusion}

\bibliographystyle{plain}
\bibliography{main}  
\input{section/7_appendix}

\end{document}

%% file: section/0_abstract.tex
\begin{abstract}
Manipulation involving rigid-deformable interactions, such as hanging clothes or dressing humans, is common in daily life, making it essential for household robots. Compared to single-object manipulation or interactions between rigid bodies, these tasks are particularly challenging due to the rich multi-point contacts and the complex dynamics of the deformable bodies during interaction. Therefore, object-centric representations such as 6D poses or structural points without task-specific information become insufficient for these interactions. In this work, we propose a hybrid correspondence-based representation tailored for rigid-deformable interactions. First, to capture intricate interaction information, we introduce structure-, task-, and interaction-aware sparse keypoints. The keypoints are generated based on the global structures of both rigid and deformable objects, and filtered by their local interaction contacts. However, tracking these sparse keypoints through the interaction remains difficult due to the high-dimensional dynamics of deformable objects. Therefore, we further construct dense correspondences on the deformable objects for accurate keypoint tracking throughout the manipulation. This hybrid design combines the advantages of both representations: sparse keypoints encode rich, task-specific information for fine-grained manipulation, while dense correspondences ensure efficient tracking and generalization to novel deformations, shapes, and scenarios. Together, they enable one-shot transfer to new tasks with minimal demonstrations. Extensive experiments demonstrate the effectiveness and broad applicability of our method. Project Page: \href{https://sparse-meets-dense.github.io/}{https://sparse-meets-dense.github.io/}.
\end{abstract}

%% file: section/1_introduction.tex
\section{INTRODUCTION}
\label{intro}
Manipulation tasks involving rigid-deformable interactions, such as hanging clothes or assisting with dressing, are common in daily-life scenarios, thus crucial for household robots. These tasks are significantly more challenging than single-object manipulation~\cite{wu2022vatmart,wu2023learningforesightfuldensevisual} or rigid-rigid interactions~\cite{li2024broadcasting} due to rich multi-point contacts and the complex, unpredictable dynamics of deformable objects~\cite{seita2021initial,blanco2024benchmarking}. Effective modeling of such interactions requires representations that capture both global object structure and local interaction contacts in a physically meaningful manner.

Existing approaches primarily rely on object-centric representations such as 6D poses~\cite{omnimanip,foundationpose} or structural keypoints~\cite{rekep,bad_points_2}. While effective for rigid objects, these representations are limited in two ways: they cannot express the contact-rich interactions in rigid-deformable tasks, nor the high-DoF nature of deformable objects \cite{behrens2019a,luo2024robot,tian2024robokeygen}. For deformable objects, recent work has shifted toward dense representations—such as affordances~\cite{wu2023learningforesightfuldensevisual,wu2025garmentpilepointlevelvisualaffordance,ha2022flingbot} or dense correspondences~\cite{wu2024unigarmentmanip,florence2018denseobjectnetslearning}. 
However, these methods face several limitations: they often require extensive annotated data, become less interpretable as object diversity increases, and introduce redundancy when applied to rigid bodies.

Therefore, we need a unified, generalizable, and data-efficient representation for such tasks.
As rigid-deformable interactions are primarily governed by object structural alignments, we propose a hybrid correspondence-based representation to represent such alignments.
We first introduce sparse keypoint correspondences (Fig.~\ref{fig:teaser}, left) to capture objects' global structures and local interaction information. To track these keypoints on high-dimensional deformable objects more accurately, we extend our representation with a dense correspondences module (Fig.~\ref{fig:teaser}, top middle). By combining the efficiency of sparse correspondence with the robustness of dense correspondences, our framework can generalize to new shapes from only a few demonstrations.

Concretely, given a single demonstration sequence and the point clouds of objects, 
we begin by extracting candidate structural keypoints on both objects. We then estimate the relative pose between the objects based on their final configuration, and use this local contact information to filter out task-relevant keypoint correspondences.

The framework then leverages the sparse keypoint correspondences to formulate constraints that can guide the robotic motion planning~\cite{rekep}. Due to the inherent alignment of keypoint pairs, they are particularly well-suited for defining such constraints to optimize manipulation trajectories.

However, constraint-based optimization requires accurate keypoint tracking over time, which is particularly challenging due to severe occlusions and the complex deformation of the soft bodies during interaction.
To address this, we introduce the dense correspondences that provide greater robustness in dynamic and continuous state spaces. Specifically, we adopt UniGarmentManip~\cite{wu2024unigarmentmanip} and extend its correspondences defined on the canonical surface of the deformable object to the 3D space. This allows us to anchor the sparse keypoints into the dense space and track them accurately during interaction.

By taking the best of sparse and dense correspondences, we extract rich task-relevant information from demonstrations to guide the manipulation. Furthermore, the framework can easily generalize to novel shapes by mapping the extracted keypoints from the demonstration to novel objects undergoing different deformations with dense correspondences. We build 4 representative deformable-rigid interaction scenarios, and experimental results showcase the superiority of our proposed framework.

In conclusion, our primary contributions are as follows:
\begin{itemize}
    \item  We study novel tasks
    with contact-rich interactions between rigid and deformable bodies, and formulate them as optimization problems with keypoint constraints;
    \item We introduce sparse correspondences for structural keypoints to extract more fine-grained task-related physical information; 
    \item We introduce dense correspondences to track the deformation of soft bodies during interaction, enabling real-time and accurate updates of constraints;
    \item Extensive experiments in both simulation and real world demonstrate the effectiveness and generalizability of our framework.
\vspace{-2mm}
\end{itemize}

%% file: section/2_related_work.tex
\section{RELATED WORK}

% \textbf{Visual Correspondences.} Learning visual correspondences aims to reflect the shared
% information (e.g., geometric, temporal and functional information) between different objects and deformations, which facilitates generalization in diverse tasks, including functional perception~\cite{lai2021functional}, 
% pose estimation~\cite{haugaard2022surfemb},
% grasping~\cite{yen2022nerf, patten2020dgcm, xue2022useek, ding2024preafford} and 
% fabric manipulation~\cite{ganapathi2021learning}. 
% While obtaining ground-truth interaction information in contact-rich object interactions is challenging, such interactions can often be represented by task-specific alignments between object structures, naturally forming correspondences that encapsulate both structural and task-relevant information.
% For deformable objects, although they have quite different geometries and deformations\cite{hopcroft1991case}, they share similar structural and topological information in the category level\cite{saha2007manipulation,sanchez2018robotic}, making correspondences a suitable representation for capturing their geometric consistency. 
% In this work, we first learn sparse spatial keypoint correspondences between the structural features of the deformable and rigid bodies. We then learn dense visual correspondences over time on the deformable body itself to enable accurate tracking of its deformations.

% \textbf{Robotic Manipulation of Rigid and Deformable objects.} 

\textbf{Structural Representations for Manipulation.} Structural representations fundamentally influence manipulation capabilities. 6D poses efficiently capture long-range object dependencies and provide occlusion robustness ~\cite{bad_pose_1, bad_pose_2, bad_pose_3,foundationpose}, but are limited to rigid objects since deformable objects lack well-defined poses. Keypoints offer flexibility and generalization ~\cite{bad_points_1,bad_points_2,bad_points_3,atm},yet are typically sparse and insufficient for capturing complex deformations of soft bodies ~\cite{yamada2024dcubed}. Moreover, keypoint proposal often requires manual task-specific annotations, limiting scalability ~\cite{chen2023keystate,xu2021an}. Although recent works ~\cite{rekep, moka, pivot} attempt to use VLMs to automate this process, they struggle to convey critical contact information for rigid-deformable interactions. To address these, we combine sparse keypoints to extract contact information while utilizing dense correspondences for deformation modeling.

\textbf{Constrained Optimization in Manipulation.} Constraints are often used to impose desired behaviors on robots~\cite{lambrecht2021optimizing}. Motion planning algorithms use geometric constraints to compute feasible trajectories that avoid obstacles and achieve goals~\cite{5152817,DBLP:journals/ijrr/SchulmanDHLABPPGA14, sundaralingam2023curoboparallelizedcollisionfreeminimumjerk}. For sequential manipulation tasks, task and motion planning (TAMP)~\cite{garrett2021integrated,kaelbling2013integrated,kaelbling2010hierarchical} is a formulated as constraint satisfaction problems~\cite{lozano2014constraint,lagriffoul2014efficiently,lagriffoul2012constraint} with continuous geometric problems as subroutines. Recent works like ReKep~\cite{rekep} further utilize VLMs to automate this process. Building on this idea, our work also employs VLMs to analyze demonstrations and decompose the task into meaningful stages. However, instead of directly relying on VLMs to generate constraints for each stage, we draw inspiration from interaction constraints~\cite{posa2014direct,mordatch2012contact,mordatch2012discovery,posa2016optimization} and leverage the physical interaction between rigid and deformable objects to derive constraints.

%% file: section/3_problem_formulation.tex
\section{PROBLEM FORMULATION}
\label{method:formulation}
Given a rigid-deformable task $\mathcal{T}$, we use VLM to decompose it into sequential stages $\mathcal{S}=\{\mathcal{S}_1,\mathcal{S}_2,\dots,\mathcal{S}_n\}$, each representing an independent sub-task. For instance, inserting a hanger into a shirt requires: $\mathcal{S}_1$=grab hanger, $\mathcal{S}_2$=lift collar, $\mathcal{S}_3$=insert right side, $\mathcal{S}_4$=insert left side, $\mathcal{S}_5$=lift up.

Our study mainly focuses on the manipulation stage containing rigid-deformable interactions (e.g., $\mathcal{S}_3$, $\mathcal{S}_4$). Given \textbf{one demonstration sequence}, the goal is to generate a successful action sequence $\mathcal{A} = \{{P}^{ee}_1, {P}^{ee}_2, \dots, {P}^{ee}_n\}$, where each ${P}^{ee}_i \in SE(3)$ represents an end-effector pose. The action sequence is considered successful when interaction regions between rigid and deformable objects align within a desired spatial threshold.

Since target interaction regions are difficult to represent directly due to occlusion and dynamics, we introduce sparse keypoint correspondences ($\mathcal{C}_{sparse}$) to capture contact information. To enable robust keypoint tracking, we employ pretrained dense correspondences ($\mathcal{C}_{dense}$) for fine-grained feature alignment on deformable surfaces. For new object pairs $(R', D')$, we use $\mathcal{C}_{dense}$ to map demonstrated sparse keypoints onto new instances, ensuring consistent contact reasoning across shape variations.

%% file: section/4_method.tex
\input{figure/framework}
\section{METHOD}
\label{method}
% Section~\ref{method:stage} first describes stage decomposition via VLMs. 
% Then, Section ~\ref{method:sparse} 
% % We begin with a brief formulation of the problem (Section ~\ref{method:formulation}). 
% describes the sparse correspondences containing structural keypoints of objects (Section ~\ref{method:resolution}) and structural keypoints of interactions (Section ~\ref{method:proposal}). Next, Section ~\ref{method:dense} covers the constraint optimization procedure with dense correspondences. Finally, Section ~\ref{method:pipeline} introduces the generalization to novel objects.
Our framework consists of four components: stage decomposition via VLMs (Section~\ref{method:stage}), sparse correspondences for structural and interaction modeling (Section~\ref{method:sparse}), constraint optimization with dense correspondences (Section~\ref{method:dense}), and generalization to novel objects (Section~\ref{method:pipeline}).

\subsection{Stage Decomposition}
\label{method:stage}

In our formulation, complex robotic tasks are decomposed into stages, which allows for
the precise definition of task requirements and facilitates the execution of complex manipulation tasks.
As shown in Fig.\ref{fig:method}, given a demonstration sequence and the task description, we first utilize VLM to decompose the task into multiple stages $\mathcal{S}$. These stages are then further categorized into two functional types: \textit{auxiliary stages}, which involve preparatory actions, and \textit{interaction stages}, which involve contact-rich interactions between rigid and deformable objects.

% The above sub-sections described how to integrate sparse and dense correspondences to efficiently model the interactions of rigid and deformable objects.
% However,
% in the real-world manipulation scenarios,
% except for the manipulation involving 
% there exist multiple manipulation stages to accomplish a task.

% In the preceding module, we provided a detailed description of how to integrate sparse and dense correspondences to represent the interactions of rigid and deformable objects.
% However, in real-world scenarios involving such interactions, a number of auxiliary operations are often required. For instance, in the task of inserting a hanger into a shirt, preparatory steps such as lifting the collar and grasping one end of the hanger are necessary before the actual interaction takes place. To effectively handle these long-horizon tasks, we propose a hierarchical manipulation pipeline.

% Specifically, for video demonstrations of these tasks, we employ vision-language models (VLMs) to decompose the demonstration into multiple stages and provide semantic descriptions for each. These stages are broadly categorized into two types: auxiliary operation stages and rigid–deformable interaction stages. The latter are addressed using the method introduced in previous sections. 
For auxiliary operation stages, the robot typically manipulates only a single object, and the operation point is often contained within the keypoints that constitute the sparse correspondences. Therefore, we construct the input prompt for VLMs using both the stage description and the candidate sparse keypoints. The VLMs then output both the specific manipulation point required for the auxiliary action and the constraint conditioned on that point.

% In the following sections, we will describe in detail how our framework handles the interaction stages, which constitute the core focus of our method.

\vspace{-2mm}
\subsection{Sparse Correspondences Modeling Interactions}
\vspace{-1.5mm}
\label{method:sparse}

In rigid-deformable manipulation tasks, obtaining accurate ground-truth interaction information is challenging due to surface deformation, complicated contact regions and dynamics. Motivated by the observation that most rigid-deformable interactions are mainly determined by the structural alignment of the target deformable and rigid objects, we propose to extract keypoints that can efficiently and effectively indicate such alignment, with the awareness of the object structures, contacts and the task. To this end, we first extract structural keypoints of objects (Section ~\ref{method:resolution}), then we further build keypoint correspondences to model interactions (Section ~\ref{method:proposal}).

% \vspace{-2.5mm}
\subsubsection{Structural Keypoints of Objects}
% \vspace{-1.5mm}
\label{method:resolution}

For both rigid and deformable objects, the structural keypoints can effectively represent them, and efficiently generalize to novel shapes.

% To capture the structural characteristics of the interacting objects, we first extract a set of structural keypoints from both the rigid and deformable objects.

For rigid objects, which often exhibit simple and stable geometric structures, we leverage its geometric and topological structure to efficiently select keypoints.
In contrast, deformable objects have more complex and dynamic structures, making it harder to extract representative keypoints by only using geometric methods. To address this, we extract structural keypoints based on their underlying skeletons, which better capture the object's topology and deformation behavior.

For a class of rigid-deformable interaction tasks, given the point clouds of the rigid and deformable bodies (denoted as $PC_r$ and $PC_d$), we first sample them to extract candidate keypoints with global structural features. Specifically, for $PC_r$, we adopt Farthest Point Sampling (FPS) to efficiently select $N_r$ candidate keypoints, forming the set $\mathcal{K}_r=\{k_r^{1},k_r^{2},\dots,k_r^{N_r}\}$ ($N_r \in [20, 100]$). For $PC_d$, we utilize Unigarmentmanip~\cite{wu2024unigarmentmanip} to extract $N_d$ skeleton points of the deformable body, forming the set $\mathcal{K}_d=\{k_d^{1},k_d^{2},\dots,k_d^{N_d}\}$ ($N_d \in [40, 200]$). These candidate keypoints are then further refined based on task-specific information, as detailed in (Section ~\ref{method:proposal}).

% \vspace{-2.5mm}
\subsubsection{Structural Keypoints of Interactions}
% \vspace{-1.5mm}
\label{method:proposal}
% After sampling an appropriate number of candidate keypoints, we will focus on \textbf{how should they be proposed}. 
While the above keypoints indicate global structural information of rigid and deformable objects, they are insufficient to capture the fine-grained local physical interaction details. Therefore, in this section, we will further consider keypoints that encapsulate local interaction information.

Given a demonstration, the final state of object interaction often contains sufficient contact information. However, due to the severe occlusions that typically occur between rigid and deformable bodies at this stage, it is challenging to extract such information directly from visual data. Therefore, we try to derive it using more robust positional information instead. 

Specifically, we manually annotate the relative pose between the rigid and soft bodies in the final state and transform $PC_r$ and $PC_d$ into the same 3D space according to the relative pose. 
Next, we traverse the keypoint sets $K_r$ and $K_d$, and establish sparse correspondences by identifying mutually nearest keypoints between the rigid and deformable objects. Specifically, for each $k_r \in K_r$, we find its nearest neighbor $k_d \in K_d$, and require that $k_r$ is also the nearest neighbor of $k_d$ in $K_r$. In addition, we impose a distance threshold $\tau$ to retain only keypoint pairs that are close enough in interaction. The resulting sparse correspondences set $\mathcal{C}$ is defined as:

\begin{equation}
  \begin{split}
  \mathcal{C} = \Big\{ (k_r^i, k_d^j) \mid &\; j = \arg\min_{j' \in \{1, \dots, N_d\}} \|k_r^i - k_d^{j'}\|, \\
  &\; i = \arg\min_{i' \in \{1, \dots, N_r\}} \|k_r^{i'} - k_d^j\|, \\
  &\; \|k_r^i - k_d^j\| < \tau \Big\}
  \end{split}
\end{equation}
% This one-to-one sparse correspondence captures the alignment of two objects in final states.

In order to execute the real-time constraint optimization more efficiently, we perform an additional point sampling operation on $\mathcal{C}$ to minimize the number of point pairs while preserving the geometric and interaction information. We apply the Furthest Point Sampling (FPS) algorithm to $K_r$ (or $K_d$) and obtain indices $I=\{i_1,i_2,…,i_{N_s}\}$ (or $I'=\{i_1',i_2',…,i_{N_s}'\}$), where $N_s (3\le N_s\le 6)$ is the number of sampled points. The final sparse correspondences $\mathcal{C}_{sparse}$ are then given by:
\begin{equation}
    \mathcal{C}_{sparse} = \left\{ \left( k_r^{{i_k}}, k_d^{{i_k}} \right) \mid i_k \in \{ i_1, i_2, \dots, i_{N_s} \} \right\}
\end{equation}

\subsection{Constraint Optimization for Manipulation Sequences}
\label{method:dense}
After obtaining the sparse correspondences, a natural question arises: how can these keypoint pairs be utilized to guide the manipulation task? In most existing works, the keypoint information is typically used as constraints to optimize the robotic motion planning. This process involves two fundamental questions: (1) how should these constraints be formulated (Section ~\ref{method:constraint_formulation}), and (2) how can these constraints be effectively optimized (Section ~\ref{method:optimization})?

\subsubsection{Optimization Constraints}
\label{method:constraint_formulation}
% While many existing works~\cite{li2025hamsterhierarchicalactionmodels,rekep,copa,voxposer} leverage large language models (LLMs) or vision-language models (VLMs) to generate constraints based on keypoint information, these models have limited spatial perception capabilities and often fail to fully utilize available visual information. Moreover, they tend to struggle with generalization when faced with novel scenarios.
% To address the aforementioned issues and more effectively utilize the extracted keypoints, we propose two types of constraints:

% mathematical formulation:
% what is the goal?
% given the goal, how to add constraints
% Ruihai: why positional and directional

To enable precise modeling of interactions between the rigid object and the deformable object, it is essential to focus on their alignment indicated by sparse keypoints. To quantify this alignment and thus guide the optimization of manipulation sequence, we decompose it into two components:
% To enable precise interaction between the rigid object and the deformable object, it is crucial to capture both positional and directional changes of the sparse keypoints. Therefore, we explicitly formulate the interaction as a combination of two types of constraints: 
a \textit{positional constraint}, which aligns the rigid objects with the deformable objects in Euclidean space, and a \textit{directional constraint}, which ensures correct relative orientation between the rigid object and the corresponding local region of the deformable object.

\textbf{Positional Constraint:} This constraint is based on the positional differences between keypoint pairs and enforces consistency in the relative positions of keypoint pairs across different frames or object states. Given $(k_r^i,k_d^i) \in \mathcal{C}_s$, we define a positional constraint as the Euclidean distance between them:
\begin{equation}
    \mathbf{c}_{pos} = \frac{1}{N_s}\sum_{(k_r^i,k_d^i)\in \mathcal{C}_{sparse}}\left\| k_r^i - k_d^i \right\|_2
\end{equation}

\textbf{Directional Constraint:} This constraint is derived by constructing planes from keypoints, computing their corresponding normal vectors, and measuring the angular difference between these normals. Since a plane is uniquely determined by three points, we use the FPS algorithm to select three representative keypoint pairs when more than three are available. These selected pairs define two planes: $P_r=\{k_r^{i_1}, k_r^{i_2}, k_r^{i_3}\}$ on rigid object and $P_d=\{k_d^{i_1}, k_d^{i_2}, k_d^{i_3}\}$ on deformable object. The normal vectors of the two planes are computed as:
% \[
% \mathbf{n}_r = \frac{(k_r^{i_2} - k_r^{i_1}) \times (k_r^{i_3} - k_r^{i_1})}
% {\left\|(k_r^{i_2} - k_r^{i_1}) \times (k_r^{i_3} - k_r^{i_1})\right\|}, \quad
% \mathbf{n}_d = \frac{(k_d^{i_2} - k_d^{i_1}) \times (k_d^{i_3} - k_d^{i_1})}
% {\left\|(k_d^{i_2} - k_d^{i_1}) \times (k_d^{i_3} - k_d^{i_1})\right\|}
% \]
\begin{equation}
  \begin{split}
  \mathbf{n}_r &= \frac{(k_r^{i_2} - k_r^{i_1}) \times (k_r^{i_3} - k_r^{i_1})}
  {\left\|(k_r^{i_2} - k_r^{i_1}) \times (k_r^{i_3} - k_r^{i_1})\right\|}, \\
  \mathbf{n}_d &= \frac{(k_d^{i_2} - k_d^{i_1}) \times (k_d^{i_3} - k_d^{i_1})}
  {\left\|(k_d^{i_2} - k_d^{i_1}) \times (k_d^{i_3} - k_d^{i_1})\right\|}
  \end{split}
\end{equation}
We then use the angular deviation to define the directional constraint as follows:
\begin{equation}
    \mathbf{c}_{ori}=\Delta \theta = \arccos(\mathbf{n}_r \cdot \mathbf{n}_d)
\end{equation}

To optimize the action sequence $\mathcal{A}$, it is necessary to establish a connection between constraint function and end-effector pose $P^{ee}$. For rigid object grasping, the target keypoint ${k}_r$ can be directly computed via a rigid transformation of $P^{ee}$:
$ {k}_r = \phi_r({P}^{ee})$
For deformable object grasping, we identify the keypoint ${k}_d$ nearest to the end-effector and estimate its position based on the current end-effector pose:
$ {k}_d = \phi_d({P}^{ee})$.

% In addition, auxiliary cost functions (e.g., collision avoidance) will be incorporated to form the final constraint, as shown in the following formula:
% \begin{equation}
%     \mathbf{c}=\lambda_1 \mathbf{c}_{pos}+\lambda_2 \mathbf{c}_{ori}+\lambda_3 \mathbf{c}_{aux} , \quad \lambda_1 + \lambda_2 + \lambda_3 = 1
% \end{equation}

% cpos cori和ee的关系

\subsubsection{Constraint Optimization Guided Manipulation with Dense Correspondences}
\label{method:optimization}

% Ruihai:
% action sequence (a1 a2 ...)

Once the interaction primitives and the corresponding spatial constraint are defined for each stage, the task execution can be formulated as a closed-loop optimization problem.
To obtain the action 
sequence \( \mathcal{A} = \{P^{ee}_t\}_{t=1}^n \), we optimize the end-effector pose at each time step $t$ by
minimizing the loss function. The optimization problem can be expressed as:
\begin{equation}
    P^{ee}_t=arg\min_{P^{ee}_t}\left\{\sum_{j=1}^{N}\mathcal{L}_j(P^{ee}_t)\right\},  \mathcal{L}=\{\mathbf{c}_{pos},\mathbf{c}_{ori}\}
\end{equation}
To enable closed-loop planning, we track keypoint pairs on both rigid and deformable objects. 
For rigid bodies, we follow prior studies~\cite{cotracker} and leverage 3D tracker to track keypoints.
For deformable objects, tracking sparse keypoints is challenging due to their high-dimensional state space and unpredictable deformations. Most existing research on deformable bodies seeks to use dense representations to capture their deformations~\cite{wu2024unigarmentmanip, florence2018denseobjectnetslearning}. 
Following~\cite{wu2024unigarmentmanip}, we define dense correspondence as a point-wise matching between two garments ($O_1$, $O_2$), which evaluates the correspondence (normalized to $[-1, 1]$) in topology or function between each point pair $(p_1, p_2)$, with $p_1$ from $O_1$ and $p_2$ from $O_2$.
% Based on the ideas of these works, we propose dense correspondence to try to solve this problem.
However, while these works build dense correspondence with deformable objects on tables or beds (for pick-and-place actions),
our task requires point-level correspondences of objects in arbitrary 3D poses.
% and we efficiently build such representations.
% Building upon UniGarmentManip~\cite{wu2024unigarmentmanip}, we extend dense correspondence from 2D cloth surfaces to a broader range of deformable objects in 3D space. 

As directly obtaining reliable 3D point-level correspondences is challenging due to the instability of soft objects, we propose a data-efficient strategy that distills pretrained correspondences knowledge from UniGarmentManip~\cite{wu2024unigarmentmanip}, which is trained on garments in canonical near-planar states (e.g., laid on a flat surface). We leverage this planar prior to initialize dense feature learning in 3D space.
Given a 3D point cloud $\mathcal{P}_{3D} \in \mathbb{R}^{N \times 3}$, we first obtain its planar projection:
% \begin{equation}
%     \mathcal{P}_{2D} = \Pi_{xy}(\mathcal{P}_{3D}),\quad \text{where } \Pi_{xy} \text{ denotes projection along the Z-axis.}
% \end{equation}

\begin{equation}
    \mathcal{P}_{2D} = \Pi_{xy}(\mathcal{P}_{3D})
\end{equation}
where $\Pi_{xy}$ denotes projection along the Z-axis.

This projection serves only to align with the pretrained planar feature extractor and does not constitute the final representation.
% We then use the pretrained 2D UniGarmentManip model to extract 2D dense features:
% \begin{equation}
%     \mathcal{F}_{2D} = \phi_{2D}(\mathcal{P}_{2D}), \quad \mathcal{F}_{2D} \in \mathbb{R}^{N \times d},\quad \text{where }\phi_{2D} \text{ is the feature extractor.} 
% \end{equation}
% where $\phi_{2D}$ is the feature extractor.

We then use the pretrained UniGarmentManip model to extract 2D dense features:
\begin{equation}
    \mathcal{F}_{2D} = \phi_{2D}(\mathcal{P}_{2D}), \quad \mathcal{F}_{2D} \in \mathbb{R}^{N \times d}
\end{equation}
where $\phi_{2D}$ is the feature extractor.

Then, we train a PointNet++ nework $\psi_{3D}$ that directly operates on the full 3D coordinates:
% To lift these 2D-derived features back into 3D, we train a PointNet++ $\psi_{3D}$ to regress them to 3D coordinates:
\begin{equation}
    \hat{\mathcal{F}}_{3D} = \psi_{3D}(\mathcal{P}_{3D}).
\end{equation}

We initialize the network by distilling the pretrained features:
\begin{equation}
    \mathcal{L}_{\text{distill}} = \left\| \hat{\mathcal{F}}_{3D} - \mathcal{F}_{2D} \right\|_2^2.
\end{equation}

To further improve the accuracy and robustness of dense correspondences, we annotate a relatively small set of 3D deformable object pairs with ground-truth correspondences $\mathcal{C}_{GT}$ and fine-tune the network with a contrastive loss $\mathcal{L}_{\text{con}}$ defined over matched and unmatched point pairs. This stage allows the network to incorporate full 3D geometric cues.
% For training details, please refer to the appendix.

% Due to the dynamics of deformable objects, 
% it is difficult to directly acquire reliable ground-truth correspondences.
% To address this, we propose a data-efficient training strategy that distills knowledge from the 2D UniGarmentManip framework into the 3D domain. Specifically, we collect a large set of 3D point clouds of deformable objects and project them onto the XY plane along the Z-axis. This projection enables us to leverage the pretrained 2D UniGarmentManip to extract features in 2D, which are then lifted back into 3D space.

% We then train a PointNet++ network to regress the 2D-derived features on the full 3D point clouds, effectively distilling 2D correspondence knowledge into the 3D domain. Finally, to further refine the model, we annotate correspondences on a relatively small subset of 3D deformable shapes and fine-tune the pretrained network accordingly.

After obtaining dense correspondences $\mathcal{C}_{dense}$, we anchor sparse keypoints to its corresponding dense representations, and use $\mathcal{C}_{dense}$ to map keypoints to their corresponding positions after deformations.

\input{figure/task}
\subsection{Generalization to Novel Objects}
\label{method:pipeline}

To further generalize this pipeline to novel objects, it is crucial to accurately transfer the $\mathcal{C}_{sparse}$ from the demonstration instances to new object pairs.
Therefore, we use $\mathcal{C}_{dense}$ to retrieve the dense feature embedding and perform a nearest-neighbor search in the dense feature space of the novel object to identify the most similar point.
Specifically, let $\mathcal{C}_{sparse}=\{(k_r^i,k_d^i)\}_{i=1}^N$ denotes sparse keypoints in demonstration, we use $\mathcal{C}_{dense}^r$ for rigid body and $\mathcal{C}_{dense}^d$ for deformable body to map $\mathcal{C}_{sparse}$ to get new $\mathcal{C}_{sparse}'$:
\begin{equation}
    \mathcal{C}_{sparse}'=\{(\mathcal{C}_{dense}^r(k_r^i),\mathcal{C}_{dense}^d(k_d^i)\}_{i=1}^N
\end{equation}
Therefore, we can use $\mathcal{C}_{sparse}'$ to formulate new constraints and optimize $\mathcal{A}'$ in new scenarios.

%% file: figure/framework.tex
\begin{strip}
\vspace{-2mm}
    \centering
    \includegraphics[width=\linewidth]{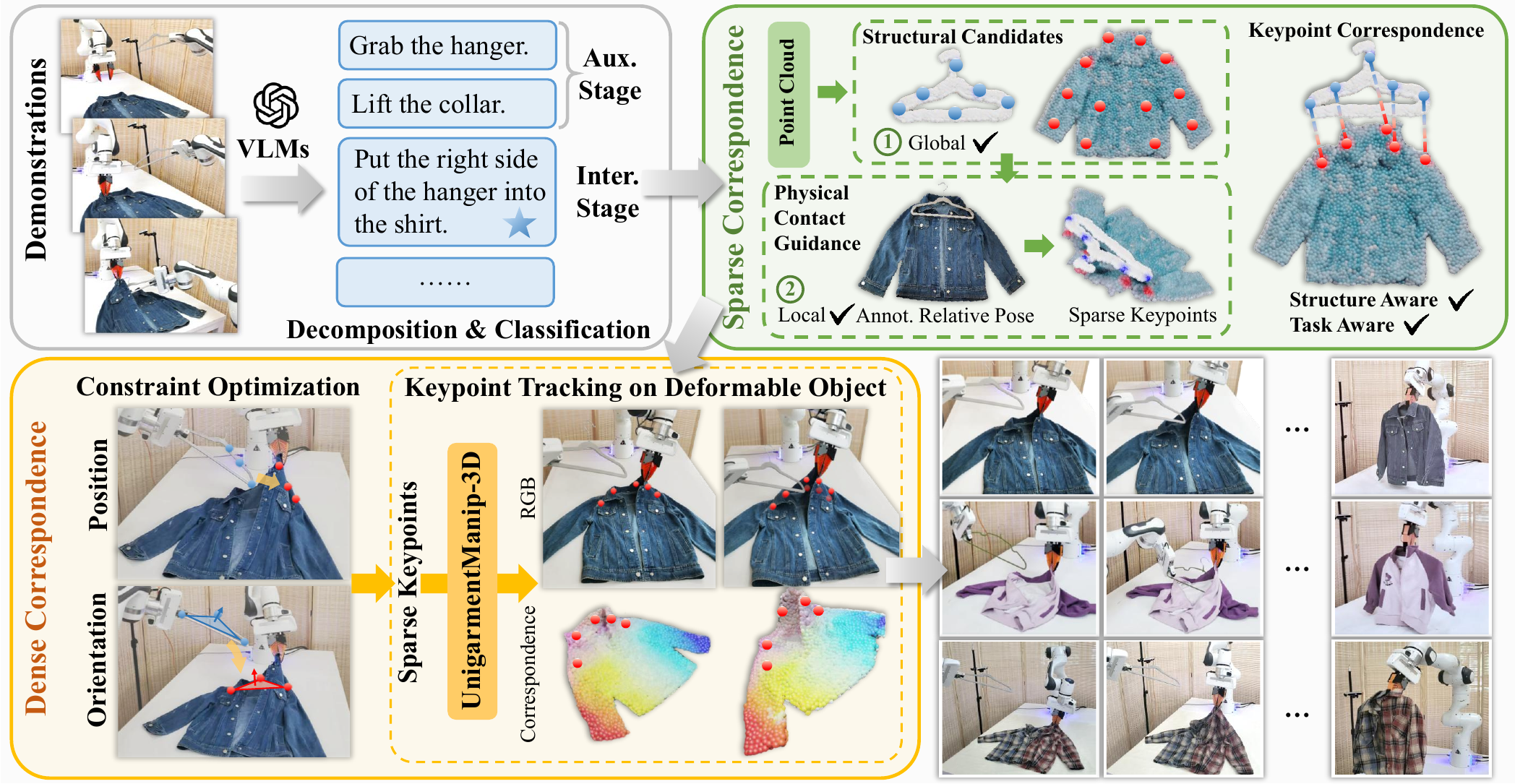}
    \captionof{figure}{\textbf{Framework Overview.} Given a task demonstration, VLMs decompose it into stages. For the stages with rigid-deformable interactions, we first extract sparse keypoints as spatial constraints, then apply constraint optimization to guide the manipulation sequence, with pretrained dense correspondence to track keypoints on deformable objects. }  
    \label{fig:method}
    \vspace{-3mm}
\end{strip}

%% file: figure/task.tex
\begin{figure}[htbp]
  \begin{center}
   \includegraphics[width=1.0\linewidth]{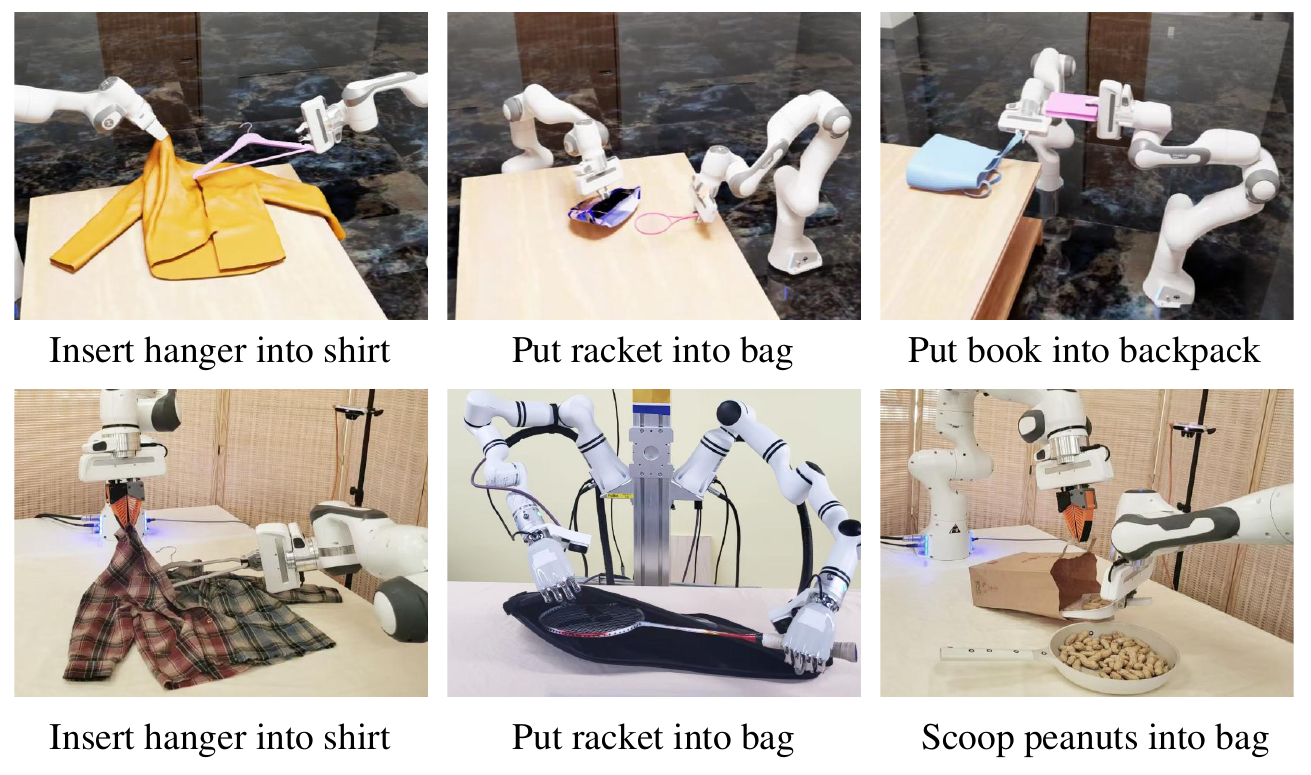}
  \end{center}
    \caption{\textbf{Task Illustrations.} Representative tasks in simulation and the real world.
  }
\label{fig:task}
\end{figure}

%% file: section/5_experiment.tex
\section{EXPERIMENT}
\label{exp}

% \vspace{-2mm}
\subsection{Environment, Assets, Data and Evaluation}
We build the multi-material simulation environment using GarmentLab~\cite{lu2024garmentlabunifiedsimulationbenchmark} implemented on Isaac Sim 4.5.0~\cite{NVIDIA_Isaac_Sim}. As shown in Fig.\ref{fig:task}, we construct 4 different representative and realistic scenes: (1) Insert the hanger into the shirt; (2) Put racket into bag; (3) Put books into backpack; (4) Scoop peanuts into bag.
For each task, we prepare a number of distinct rigid and deformable object assets with varying shapes to ensure diversity and generalization, and their configurations are generated automatically through code.

% \vspace{-3mm}
\subsection{Baselines}
% \vspace{-1.5mm}
We evaluate our method in both simulation and real-world scenarios.  We compare our approach with two baselines:
(1) \textbf{Annot. Rekep}~\cite{rekep}, which employs human-annotated relational keypoint constraints and hierarchical optimization for real-time action generation; (2) \textbf{UniGarment}~\cite{wu2024unigarmentmanip}, which guides constraint formulation and optimization by establishing dense correspondences between rigid and deformable objects. Among them, Rekep relies entirely on sparse representations, while UniGarment adopts a fully dense representation. We compare our approach against both to highlight the advantages of taking the best of sparse and dense correspondences.

\subsection{Results and Analysis}
\label{results}
\input{table/baseline_table}
Tab.~\ref{tab:baseline} shows quantitative comparisons with baselines. 
While Annot. Rekep utilizes manually annotated keypoints for its constraint optimization, it still underperform compared to our method. The main reason lies in its reliance on DINOv2 feature matching when tracking keypoints.
% Annot. Rekep performs poorly on rigid-deformable interaction tasks due to two main reasons.
% First, it relies on DINOv2 features to select keypoints. However, as shown in the first row of Fig.~\ref{fig:rekep}, DINOv2 fails to capture the structural features of deformable objects, let alone interaction-relevant features which are crucial for intricate rigid-deformable manipulation. As a result, the selected keypoints often fail to guide the interaction correctly.
% Second, Rekep tracks keypoints via DINOv2 feature matching. 
As illustrated in the second row of Fig.~\ref{fig:rekep}, DINOv2 features lack consistency under soft-body deformation, leading to tracking errors and ultimately constraint optimization failures.
UniGarment uses dense correspondences between rigid and deformable objects, providing useful interaction cues for constraint formulation. However, the high density increases computational cost, and inconsistent matches in occluded or highly deformable regions can reduce overall accuracy.

\subsection{Ablation Study}

\input{table/ablation_table}
To demonstrate the necessity of the proposed correspondences as well as designed constraints, we compare with the following ablated versions:  
(1) Ours w/o $\mathcal{C}_{sparse}$ that uses DINOv2 to generate keypoints instead of $\mathcal{C}_{sparse}$;
(2) Ours w/o $\mathcal{C}_{dense}$ that tracks keypoints by matching pixel-wise DINOv2 features across frames instead of using $\mathcal{C}_{dense}$;
(3) Ours w/o $c_{pos}$ that removes the positional constraint;
(4) Ours w/o $c_{ori}$ that removes the orientational constraint;
(5) Ours w/ ${3D-Tracker}$ that uses SpatialTracker~\cite{xiao2024spatialtrackertracking2dpixels} to track keypoints.
Tab.\ref{tab:compare} shows that sparse correspondences, composed of keypoints aware of structures and interactions, plays a crucial role in manipulation. Dense correspondences are also important. 
Compared to feature tracking based on the DINOv2 or recent 3D trackers, dense correspondence performs better in scenes with complex dynamics, where accurate tracking of deformable-object keypoints is required.
The positional constraint is crucial for nearly all tasks. In contrast, the orientation constraint can be omitted in a few interactions where precise alignment is not strictly required.
\input{figure/ablation}
\input{figure/generalization}
\subsection{Generalization}

\begin{table}
    \centering
    \begin{tabular}{lccc}
    \toprule
    \textbf{Task}    & \textbf{ReKep} & \textbf{UniGarment} &  \textbf{Ours} \\ \midrule 
    Insert hanger into shirt  &15.6\% & 57.1\%   & \textbf{65.2}\%      \\
    Put racket into bag &20.7\% &59.3\% & \textbf{66.5}\%         \\ 
    Put books into backpack &20.9\% & 39.4\% & \textbf{69.1}\%  \\
    \bottomrule
    \end{tabular}
    \caption{\textbf{Generalization Results.} Section~\ref{results} provides detailed analysis.}
    \label{tab:generalize}
    \vspace{-12mm}
\end{table}

Tab.~\ref{tab:generalize} and Fig.~\ref{fig:generalization} demonstrate the generalization capability. For each task, we use 1 high-quality demonstration to extract sparse keypoints, which are mapped to novel objects with dense correspondences. Our generalization covers variations in object shapes, deformations and poses.

\subsection{Real-World Evaluation}
For real-world evaluation, we use a Franka Panda robot and two RealMan RM75-6F arms, each equipped with a Psibot G0-R dexterous hand. Depth images are captured using RealSense cameras. The last colum of Tab.~\ref{tab:baseline} reports success rates. Supplementary video shows more demonstrations.

%% file: table/baseline_table.tex
% \begin{table}[ht]
%     \centering
% \setlength\tabcolsep{2.5pt}%调列距
%     \begin{tabular}{lcccc}
%     \toprule
%     &   Wholly Sparse &Wholly Dense& &\\ 
%     \cline{2-2} \cline{3-3}
%     Tasks (Simulation)   &Annot. ReKep&UniGarment &Ours\\ 

%      \midrule \midrule  
%     Insert hanger into shirt                     & 60.9\% & 62.3\%  & \textbf{71.4\%}   \\
%     Put racket into bag    & 65.2\% & 65.4\%  & \textbf{73.6\%}  \\   
%     Put books into backpack        & 58.7\% & 46.8\%  & \textbf{78.2\%}  \\   
%     Scoop peanuts into bag & 49.1\% & 45.1\% & \textbf{70.9\%}\\  
%      \midrule 
%     Total & 58.5\% & 54.9\%  & \textbf{73.5\%} \\ 
%     \midrule
%     &   Wholly Sparse &Wholly Dense& &\\ 
%     \cline{2-2} \cline{3-3}
%     Tasks (Real)   &Annot. ReKep&UniGarment &Ours\\ 
%      \midrule  \midrule 
%     Insert hanger into shirt                     & 5/10 & 5/10  & \textbf{7/10}  \\
%     Put racket into bag    & 5/10 & 6/10  & \textbf{6/10}   \\   
%     Put books into backpack        & 5/10 & 5/10  & \textbf{7/10}  \\   
%     Scoop peanuts into bag & 4/10 & 4/10 &  \textbf{6/10}  \\   
%     \midrule
%     Total & 47.5\% & 50\%  & \textbf{65.0\%}  \\
%     \bottomrule
%     \end{tabular}
%     \vspace{2mm}
%     \caption{\textbf{Quantitative Comparisons with Baselines.} Our method outperforms baselines by a margin.}
%     \label{tab:baseline}
% \end{table}

\begin{table}[ht]
    \centering
\setlength\tabcolsep{2.8pt}
    \begin{tabular}{l*{3}{c}}
    \toprule
    & \multicolumn{2}{c}{\textbf{Baseline Methods}} & \\
    \cmidrule(lr){2-3}
    \textbf{Tasks} & \textbf{Wholly Sparse} & \textbf{Wholly Dense} & \textbf{Ours} \\
    & \textit{(ReKep~\cite{rekep})} & \textit{(UniGarment~\cite{wu2024unigarmentmanip})} & \\
    \midrule
    \multicolumn{4}{c}{\textsc{Simulation Results}} \\
    \midrule
    Insert hanger into shirt & 60.9\% & 62.3\% & \textbf{71.4\%} \\
    Put racket into bag & 65.2\% & 65.4\% & \textbf{73.6\%} \\
    Put books into backpack & 58.7\% & 46.8\% & \textbf{78.2\%} \\
    Scoop peanuts into bag & 49.1\% & 45.1\% & \textbf{70.9\%} \\
    \midrule
    \textbf{Average} & 58.5\% & 54.9\% & \textbf{73.5\%} \\
    \midrule
    \multicolumn{4}{c}{\textsc{Real-World Results}} \\
    \midrule
    Insert hanger into shirt & 5/10 & 5/10 & \textbf{7/10} \\
    Put racket into bag & 5/10 & 6/10 & \textbf{6/10} \\
    Put books into backpack & 5/10 & 5/10 & \textbf{7/10} \\
    Scoop peanuts into bag & 4/10 & 4/10 & \textbf{6/10} \\
    \midrule
    \textbf{Average} & 47.5\% & 50.0\% & \textbf{65.0\%} \\
    \bottomrule
    \end{tabular}
\caption{\textbf{Quantitative Comparison with Baseline Methods.} Our approach demonstrates consistent improvements across all tasks in both simulation and real-world settings.}
\vspace{-2mm}
    \label{tab:baseline}
\end{table}

%% file: table/ablation_table.tex
\begin{table}[ht]
    \centering
\setlength\tabcolsep{2.5pt}%调列距
    \begin{tabular}{lcc}
    \toprule
          & Insert hanger into shirt & Put racket into bag \\ 
        \midrule
    {Ours w/o $\mathcal{C}_{sparse}$}    & 20.1\%     &   28.6\% \\
    {Ours w/o $\mathcal{C}_{dense}$} & 56.4\%   & 63.2\%  \\ 
    {Ours w/o $c_{pos}$} & 0.0\%   & 0.0\%  \\ 
    {Ours w/o $c_{ori}$} & 28.3\%   & 32.7\%  \\ 
    {Ours w/ ${3D-Tracker}$} & 56.4\%   & 63.2\%  \\ 
    {\textbf{Ours}} & \textbf{71.4\%}   & \textbf{73.6\%}  \\ 
    \bottomrule
    \end{tabular}
    \caption{\textbf{Ablation Studies.} Section~\ref{results} provides detailed analysis.}
    \label{tab:compare}
\end{table}

%% file: figure/ablation.tex
\begin{strip}
\vspace{-2mm}
    \centering
    \includegraphics[width=\linewidth]{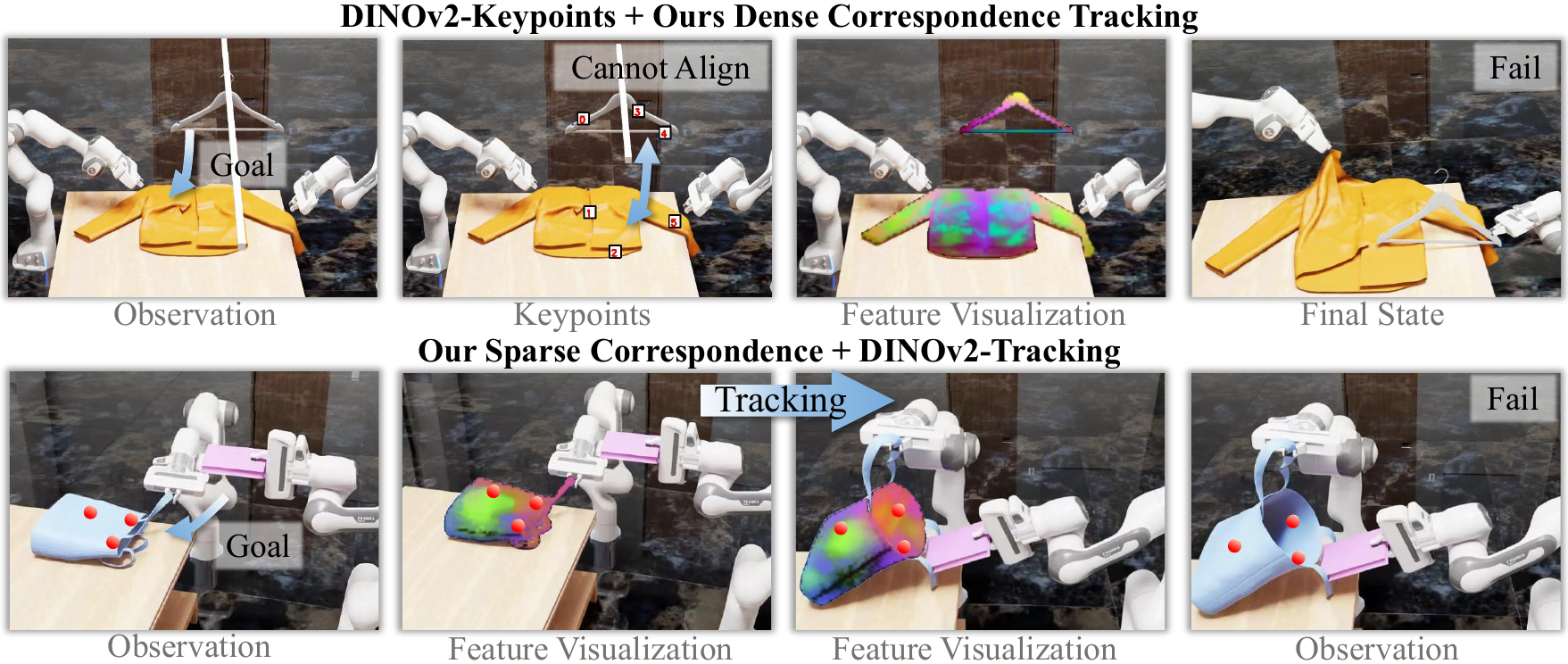}
    \captionof{figure}{\textbf{Ablation Studies.} 
    % The top and bottom rows show results of replacing $\mathcal{C}_{sparse}$ and $\mathcal{C}_{dense}$ with DINOv2, respectively.
    The top and bottom rows respectively show  results of replacing  $\mathcal{C}_{sparse}$ and $\mathcal{C}_{dense}$ with DINOv2.
    } 
    \label{fig:rekep}
\end{strip}

%% file: figure/generalization.tex
\begin{strip}
    \centering
\includegraphics[width=\linewidth]{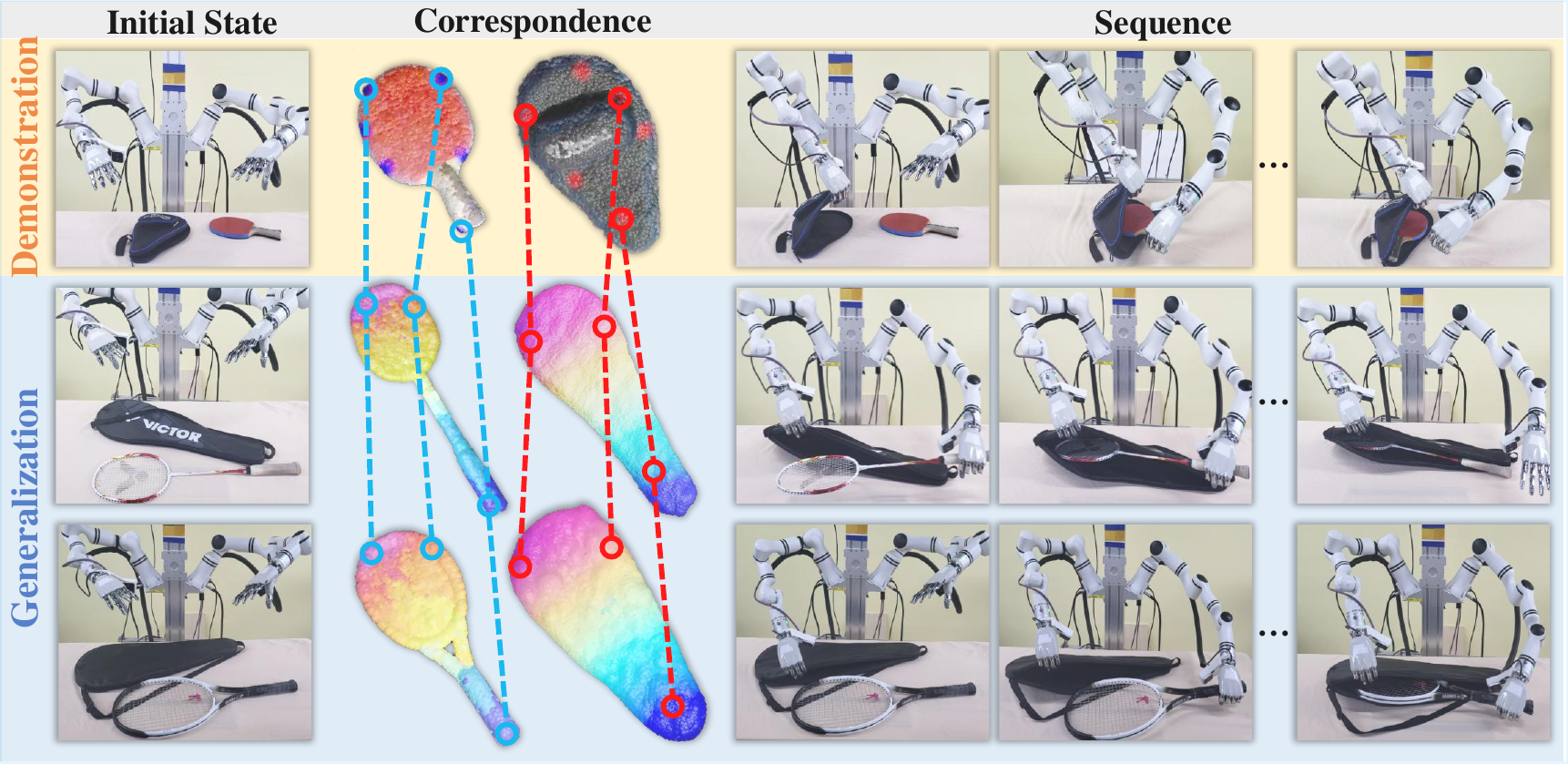}
    \captionof{figure}{\textbf{Dense Correspondences Enable Generalization.} Given the demonstration of table tennis racket (row 1), we can generalize to other rackets by mapping keypoints via $\mathcal{C}_{dense}$ (row 2,3).} 
    \label{fig:generalization}
\end{strip}

%% file: section/6_conclusion.tex
\section{CONCLUSIONS}

\label{limitation}

We present a hybrid correspondence-based representation that effectively models complex interactions between rigid and deformable objects. We first introduce structure- and interaction-aware sparse keypoints correspondences, and then dense correspondences for accurate tracking. By taking the best of both, our approach enables robust and generalizable manipulation across new shapes and deformations. Extensive experiments in varied scenarios validate the effectiveness of our method.

\textbf{Limitations and Future Work.} While our method effectively models rigid-deformable interactions, tasks with deformable-to-deformable interactions remain challenging. Defining efficient representations and modeling such interactions are difficult due to complex dynamics, entanglement and occlusion. 
% Modeling such interactions is still an open problem.

%% file: section/7_appendix.tex
\section*{APPENDIX}

\subsection{Real-world Environment}
\begin{figure*}[t!]
  \begin{center}
   \includegraphics[width=0.8\linewidth]{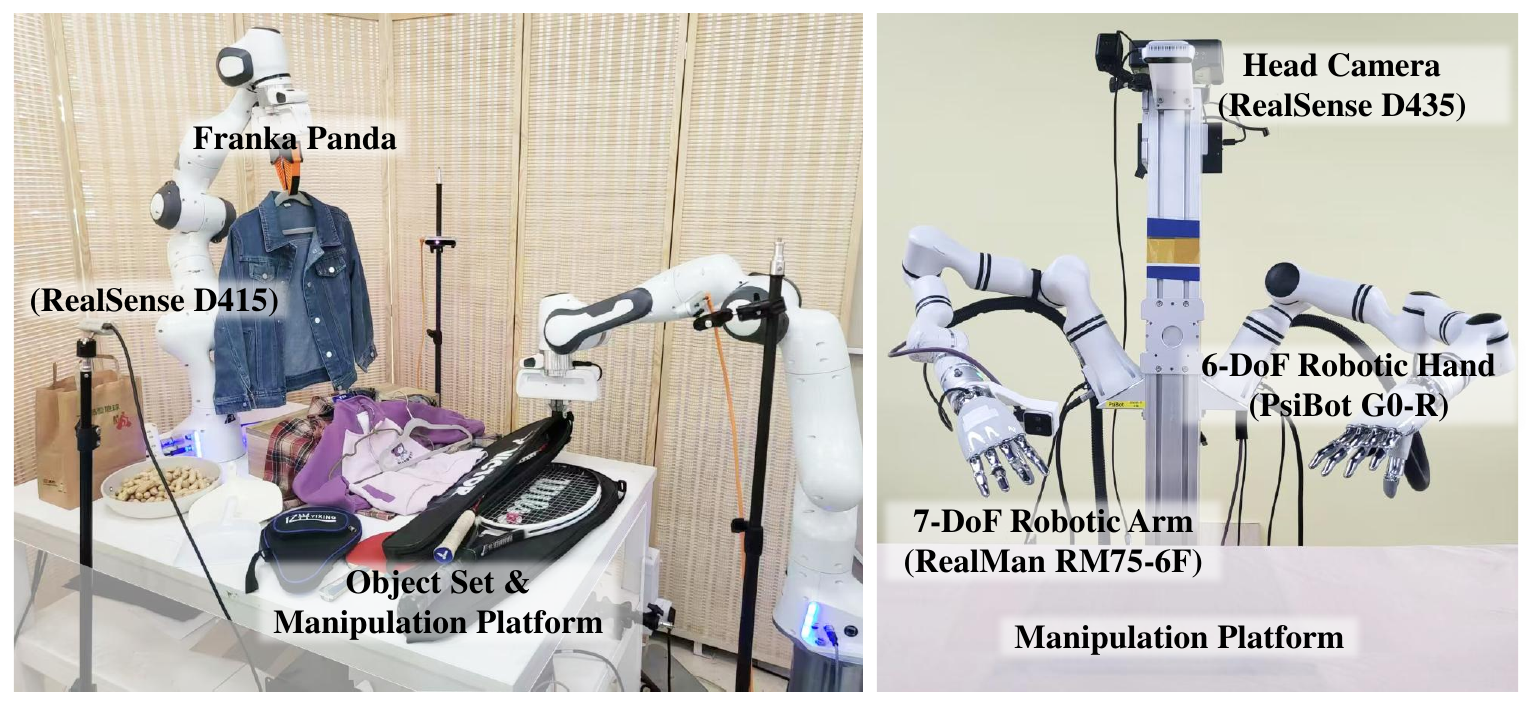}
  \end{center}
  \vspace{-3mm}\caption{
  \textbf{Real-world Environment and Assets Utilized.} 
  }
  % \vspace{-3mm}
\label{fig:asset}
\end{figure*}

% used to describe our assets and configuration, please follow Omnimanip.
%\label{app:simenv}
%\begin{figure*}[!ht]
%    \centering
%\includegraphics[width=1.0\textwidth]{img/real assets.pdf} 
%\caption{Visualizations of simulation environments.}
%\label{fig:real assets} 
%\end{figure*}
We detailed our real-world environment setup in Fig.~\ref{fig:asset}. 
% robot
Our real-world implementation is deployed on two distinct hardware platforms.

The first platform consists of a 6-DOF robotic hand (PsiBot G0-R) and a 7-DOF robotic arm (RealMan RM75-6F), both designed for dexterous manipulation tasks. For perception, three Intel RealSense D415 RGB-D cameras are deployed: one mounted at the top rear as a head camera, and the other two positioned at the front, one on the left and one on the right, providing comprehensive spatial coverage. This multi-view setup enhances the system’s ability to accurately localize objects and execute precise operations. This setup is particularly well-suited for manipulating large and heavy objects.

The second platform includes two Franka Emika Panda robotic arms, renowned for their precision and torque-controlled movements. To handle deformable objects with diverse shapes, the robot uses a standard parallel gripper equipped with UMI fingers, which provide flexibility and adaptability during grasping. For perception, three Intel RealSense D415 RGB-D cameras are positioned at the left, right, and front, providing multi-view depth and color information. 
% The entire system is powered by a high-performance workstation equipped with an Intel Core i7 processor, 64 GB of RAM, and an NVIDIA RTX 4090 GPU, ensuring real-time inference and motion planning. 

Both of the two platforms are powered by a high-performance workstation equipped with an Intel Core i7 processor, 64 GB of RAM, and an NVIDIA RTX 4090 GPU, supporting real-time inference and motion planning.

% assets
Our real-world assets are shown in Fig.~\ref{fig:asset} (left). We select multiple different instances for each task to demonstrate the generalizability of our pipeline.

\subsection{Training Details of 3D-UnigarmentManip}

In this section, we discuss the training details of 3D-UnigarmentManip. 

As described in ~\ref{method:dense}, directly obtaining reliable 3D point-level correspondences is challenging due to the instability of soft objects.
Therefore, we propose a data-efficient strategy that distills pretrained 2D correspondences knowledge from UniGarmentManip~\cite{wu2024unigarmentmanip} into the 3D space.

First,
Given a 3D point cloud $\mathcal{P}_{3D} \in \mathbb{R}^{N \times 3}$, we project it orthographically onto the XY-plane to obtain the 2D counterpart:
\begin{equation}
\mathcal{P}_{2D} = \Pi_{xy}(\mathcal{P}_{3D})
\end{equation}
where $\Pi_{xy}$ denotes projection along the $z$-axis.

We then use the pretrained 2D UniGarmentManip model to extract 2D dense features:
\begin{equation}
\mathcal{F}_{2D} = \phi_{2D}(\mathcal{P}_{2D}), 
\quad \mathcal{F}_{2D} \in \mathbb{R}^{N \times d}
\end{equation}
where $\phi_{2D}$ is the feature extractor.
% where $\phi_{2D}$ is the feature extractor.

Then, we train a PointNet++ network $\psi_{3D}$ to lift these 2D-derived features back into 3D space:
% To lift these 2D-derived features back into 3D, we train a PointNet++ $\psi_{3D}$ to regress them to 3D coordinates:
\begin{equation}
    \hat{\mathcal{F}}_{3D} = \psi_{3D}(\mathcal{P}_{3D}).
\end{equation}

We supervise the network with the feature difference:
\begin{equation}
    \mathcal{L}_{\text{distill}} = \left\| \hat{\mathcal{F}}_{3D} - \mathcal{F}_{2D} \right\|_2^2.
\end{equation}

% Finally, to further improve the robustness of dense correspondences, we annotate a relatively small set of 3D deformable object pairs with ground-truth correspondences $\mathcal{C}_{GT}$ and fine-tune the network with a contrastive loss $\mathcal{L}_{\text{sup}}$ defined over matched and unmatched point pairs. 
Finally, to further improve the robustness of dense correspondences, we annotate a relatively small set of 3D deformable object pairs with ground-truth correspondences $\mathcal{C}_{GT}$ and 
adopt the training strategy proposed in UnigarmentManip~\cite{wu2024unigarmentmanip} to
fine-tune the network. The following offers a brief overview, for more details, please refer to the original paper. 

\paragraph{Cross-Deformation Correspondences.} 
Given two observations $O$ and $O'$ of the \textit{same} garment in different deformations, and a visible correspondence $p \leftrightarrow p'$, we extract features via a backbone network $\mathbf{F}$ such that $f_p = \mathbf{F}(p)$. A contrastive loss encourages deformation-invariant features:

\begin{equation}
\mathcal{L}_{CD} = -\log \left( \frac{\exp(f_p \cdot f_{p'}/\tau)}{\sum_{i=1}^{m} \exp(f_p \cdot f_{p_i'}/\tau)} \right),
\end{equation}

where $\{p_i'\}_{i=1}^m$ are negative samples in $O'$ and $\tau$ is a temperature hyperparameter.

\paragraph{Cross-Object Correspondences.}
For different garments in canonical poses, we generate a skeleton with $S$ keypoints $\{s_i\}$ using Skeleton Merger~\cite{shi2021skeleton}.
As skeleton points are ordered,
given observation $O$ of a garment with one of its skeleton point $p$,
we can get the corresponding skeleton point $\Tilde{p}$ on the observation $\Tilde{O}$ of another garment.
Then, the topological correspondences have been built in the skeleton-point level.
To obtain point-level correspondences, given an arbitrary point $p$, we use its nearby skeleton points to reflect its feature (like interpolation).
Therefore,
the representation of each point on the garment will reflect its topology,
and  dense correspondences between different garments has been naturally built.

\begin{table*}[t!]
% \vspace{-2mm}
\centering
\resizebox{0.8\linewidth}{!}{
\begin{tabular}{lccc}
\toprule
& \textbf{Insert hanger into shirt} & \textbf{Put racket into bag} & \textbf{Put books into backpack} \\
\midrule \midrule
VLM  &25.6\% & 29.1\% & 30.2\% \\
Ours &71.4\% &73.6\% &78.2\% \\
\bottomrule
\end{tabular}
}
\caption{\textbf{Generalization Results.} Section~\ref{results} provides detailed analysis.}
\label{tab:constraints}
\vspace{-4mm}
\end{table*}

\paragraph{Integration of Cross-Deformation and Cross-Object Correspondence.}
We have established dense correspondences for the same garment under different deformations and for different flat garments. Next, we unify these into a dense representation applicable to diverse garments in any deformation.
By projecting skeletons to deformed states via point tracing, given a skeleton point \( p \) on an observation \( O \) in deformation, we find its counterpart \(\tilde{p}\) on another garment \(\tilde{O}\). If \(\tilde{p}\) is visible, their features \( f_p \) and \( f_{\tilde{p}} \) should match. Using \(\tilde{p}\) as a positive and sampling \(m=150\) negatives \(\tilde{p}_i^\prime\), we train with InfoNCE loss:
\[
\mathcal{L}_{CO} = -\log \frac{\exp(f_p \cdot f_{\tilde{p}}/\tau)}{\sum_{i=1}^{m} \exp(f_p \cdot f_{\tilde{p}_i^\prime}/\tau)}
\]
Simultaneously, we reinforce deformation-agnostic cross-object correspondence by jointly training \(\mathcal{L}_{CO}\) with the cross-deformation loss \(\mathcal{L}_{CD}\) introduced before.

\paragraph{Coarse-to-Fine Refinement.}
To better handle challenging regions, we introduce a Coarse-to-Fine (C2F) Refinement module to refine the offline-trained model based on its own online prediction failures.
Given a point $p$ on observation $O$, we predict its correspondence $p^\prime$ on $O^\prime$. We then identify a set of hard negatives $\{p_i^\prime\}$ with high correspondence scores (above $\alpha$) but large distances (above $\beta$) to $p^\prime$ in the canonical space. These false positives indicate unreliable predictions.
To emphasize these hard cases, we propose a distance-weighted contrastive loss:
\begin{equation}
\label{eq:l_c2f}
   \mathcal{L}_{C2F} = -\log\left(\frac{\exp(f_p \cdot f_{p^\prime} / \tau)}{\sum_{i=1}^{r} d_{p_i^\prime} \cdot \exp(f_p \cdot f_{p_i^\prime} / \tau)}\right)
\end{equation}
We jointly optimize $\mathcal{L}_{C2F}$ with offline losses $\mathcal{L}_{CO}$ and $\mathcal{L}_{CD}$ to refine the model while preserving general correspondence knowledge.

% \paragraph{Training Strategy.} 

\subsection{Evaluation Details}
Below we discuss the evaluation details for the experiments.
\subsubsection{Task Details}
% please refer to rekep
% success rate should emphasize "which part and which part is aligned"

We have implemented the following tasks in real-world and simulation, as shown in Fig.~\ref{fig:task}:

\textbf{Insert Hanger into Shirt}. A shirt is laid flat on the table, while a hanger is suspended from a rod. Robot $A$ first grasps the hanger and begins inserting it into the shirt after Robot $B$ lifts the collar. Then, Robot $B$ grabs the hanger and continues the insertion process while Robot $A$ lifts the other side of the collar. The task is considered successful if the hanger is fully inserted into the shirt, with its final orientation meeting predefined angular constraints.

\textbf{Insert Racket into Bag}. A racket and a racket bag are placed flat on the table. Robot $A$ should lift one side of the bag and robot $B$ must pick up the racket and precisely insert it into the bag. The task is considered successful if the racket is fully placed inside the racket bag and is geometrically aligned with it.

\textbf{Insert Book into Backpack}. A book and a backpack are laid flat on the table. Robot $A$ should lift one side of the bag and robot $B$ is required to pick up the book and insert it into the backpack. The task is considered successful if the book is placed inside the bag with correct spatial alignment.

\textbf{Scoop Peanuts into Bag}. A bag and a pile of peanuts are placed on the table. The robot must sequentially scoop the peanuts, lift the bag and place the peanuts into the bag. The task is considered successful if all peanuts are completely transferred into the bag.

For each task performed in simulation, the object configurations are generated automatically through code. In contrast, for real-world experiments, the objects are manually arranged into different configurations, after which the robots execute the corresponding procedures via programmatic control.

\subsubsection{Details on Baseline Methods}
% mainly used to introduce Unigarmentmanip. how we change this method into a baseline method.
% can refer to rekep
We adopt Rekep and UnigarmentManip as baseline methods, as both utilize point-based representations, making them well-suited for handling interactions involving rigid and deformable objects.

To ensure a fair comparison, we adapt Rekep to our setting with several modifications. Specifically, we enhance the original prompt from the paper by incorporating the prompt used in our method to provide richer contextual information. For deformable object simulation, we employ GarmentLab~\cite{lu2024garmentlabunifiedsimulationbenchmark} as the underlying environment.

UnigarmentManip is used to establish correspondences between garments in different shapes and deformation states. We extend it to compute correspondences between rigid and deformable objects, based on their relative positions in the final configuration. These dense correspondences are then used to define constraints for guiding the interaction.

\begin{figure*}[t!]
  \begin{center}
   \includegraphics[width=0.8\linewidth]{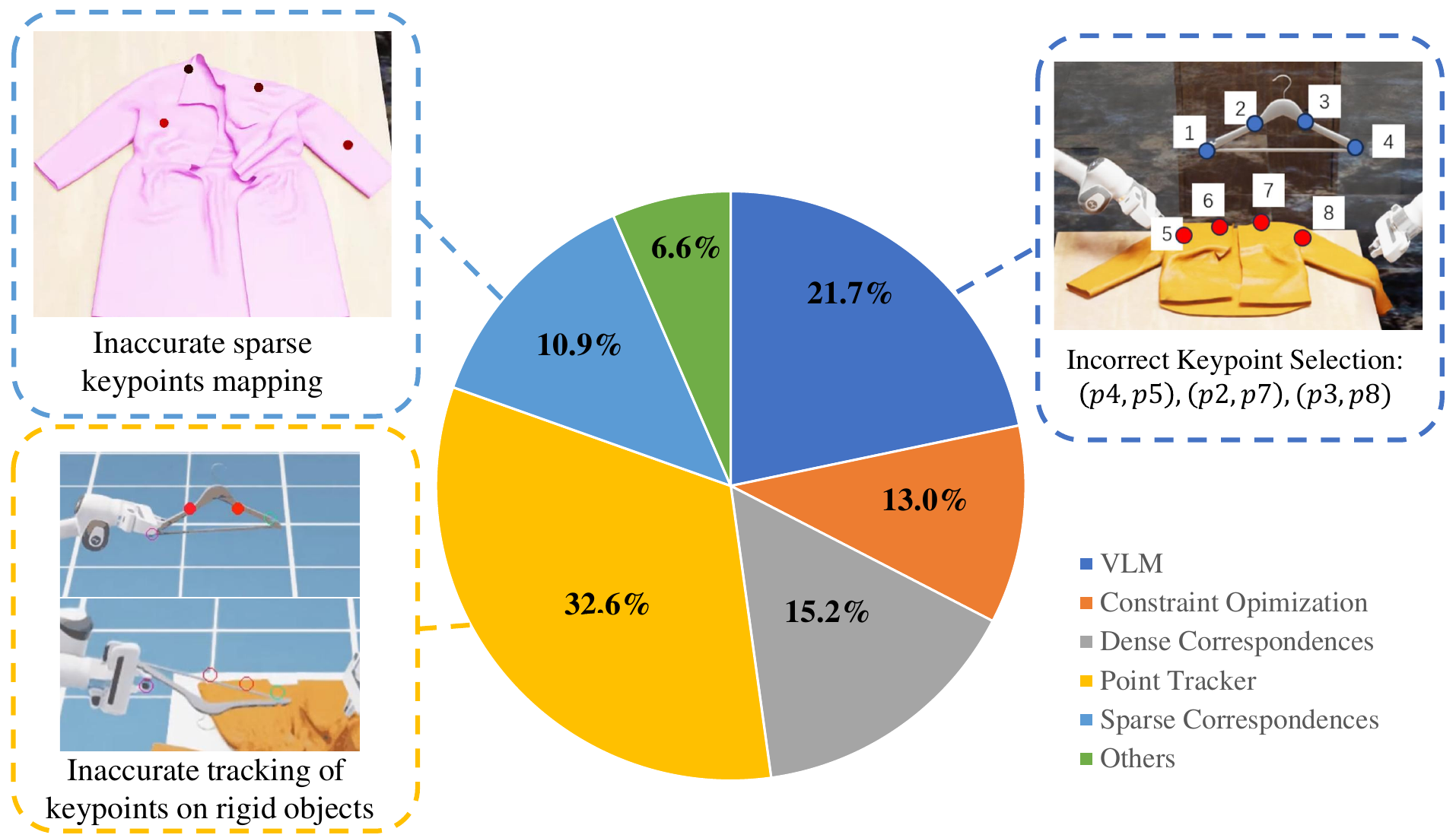}
  \end{center}
  \caption{
  \textbf{Error breakdown of
the system modules.} 
  }
  % \vspace{-3mm}
\label{fig:error}
\end{figure*}

\subsubsection{Comparison with Vision-Language Model Generated Constraints}

Many existing works~\cite{rekep,copa,voxposer} utilize Vision-Language Models (VLMs) to generate constraints for motion planning. 
However, in tasks involving rigid-deformable interactions, it is often necessary to fully exploit the correspondences between keypoints to generate precise constraints, which is a capability that current VLMs still lack.
Therefore, we conduct comparative experiments in three scenarios between constraints generated by VLMs and those designed by our method. The results are shown in the Tab.~\ref{tab:constraints}.

% \subsection{Implementation Details of Keypoint Proposal}
% % if we can find a way to get relative pose (e.g, genpose++) automatically, then we can put it here. or we can only say, we manually get this pose.
% To extract keypoints, structural alignment between the objects is required. Achieving such alignment relies on estimating their relative pose, which provides a more robust representation, especially under occlusions.

% We utilize \textbf{GenPose++} to automatically estimate the relative pose between rigid and deformable objects. This not only simplifies the pipeline but also improves the generalization ability across different scenarios. With the relative pose obtained, we follow the procedure described in the main paper to extract keypoints and compute the necessary constraints.

\subsection{Querying Vision-Language Model}
% refer to rekep and omnimanip
Since VLMs struggle to precisely analyze complex rigid-deformable interactions, we utilize them only for high-level planning. For low-level execution, we rely on our sparse correspondences to generate constraints.

Specifically, the VLMs are responsible for:
\begin{itemize}
    \item Decomposing the task into independent stages;
    \item Classifying each stage as either an \textit{interaction stage} or an \textit{auxiliary stage};
    \item Selecting the required keypoints for operation within each stage.
\end{itemize}

Given an rgb sequence, an image annotated with sparse keypoints and a textual task description, the VLM first decompose the task into individual stages. Then the VLM classifies each stage based on whether it directly involves rigid-deformable interactions (\textbf{interaction stage}) or serves as a preparatory step (\textbf{auxiliary stage}). For each stage, the VLM further proposes keypoints from the candidate sparse keypoints that may be used to construct constraints in this stage, thereby providing high-level guidance for the robot's motion.

The prompt used for the VLM is shown below.
% \begin{tcolorbox}[colback=gray!5!white, colframe=gray!75!black,
%   title=Prompt Code, fonttitle=\bfseries,
%   listing only, listing options={language=Markdown, breaklines=true, basicstyle=\ttfamily\small}]
\begin{lstlisting}[numbers=none]
## Instructions
You are given two inputs:
A single image, containing both rigid and soft objects, with precisely annotated keypoints.
A video, showing a successful execution of a similar task by a robot.

Your Objective

You must analyze the current image and generate a step-by-step manipulation plan by learning from the video.

- Learn the action sequence, logic, and manipulation flow from the video.
- Mimic the stage order and type of operations from the video.
- Apply that same strategy to the current image, using only its visible colored dot keypoints.

Keypoint Annotation Rules (STRICTLY ENFORCED)

- Every keypoint is a colored dot placed precisely on the surface of an object.
- Red dots indicate keypoints on soft objects.
- Blue dots indicate keypoints on rigid objects.
- Each dot is labeled with a numeric ID (e.g., 1, 2, 3...), shown clearly next to the dot in the image.
- Use only the keypoints that appear as colored dots in the image.
- Do NOT invent, rename, skip, or hallucinate keypoints.
- The index number next to each colored dot is the ONLY valid reference for that keypoint.
- If a keypoint is not explicitly shown as a colored dot with a number, you must NOT refer to it in any reasoning, operation, or constraint.

Robot Context

- The robot primarily manipulates rigid objects.
- It only interacts with soft objects when necessary to assist rigid object manipulation (e.g., lifting fabric to enable hanger insertion).

Step 1: Task Inference and Decomposition

From visual cues in the image, infer the most likely manipulation task.

Then, break the task into clear, sequential robot actions. Each step must:

- Represent one robot action only.
- Treat grasping as a separate, explicit stage.
- Explain why each action is needed (its purpose).

Example:  
Task: "Insert hanger into shirt"  
- Stage 1: Grasp the hanger (to gain control of it)  
- Stage 2: Lift collar of shirt (to make space for hanger)  
- Stage 3: Insert hanger into shirt (achieve final configuration)

Step 2: Stage Classification

Classify each stage as:

- Auxiliary Stage: Prep action (e.g., lifting, opening) without direct soft and rigid object interaction.  
  Example: grasp the hanger; lift the collar
- Interaction Stage: Involves direct contact or alignment between rigid and soft objects  
  Example: insert the hanger into the shirt; insert the book into the bag

Step 3: Keypoint Selection and Constraints

You must only refer to valid, colored-dot keypoints as labeled in the image.

FOR AUXILIARY STAGES:

- Select ONE keypoint on the object.
- Specify the operation (e.g., lift keypoint 3, grasp keypoint 5).

FOR INTERACTION STAGES:

- Select pairs of keypoints across objects.
- Each pair of keypoints must be the colored dots from the image.
- Define:
  - The operation (e.g., insert, align, place)
  - The constraint, e.g., (4,7) = keypoints 4 and 7 should come together.

Valid Example:  
Selected Keypoints: 3, 6  
Operation: align rigid arm at 3 with cloth point 6  
Constraint: (3,6)

Invalid Example:  
Selected Keypoints: "end of the sleeve", "hanger tip": Not using numbered keypoints!

Step 4: Output Format (Use This Template)

- Inferred Task:  
(Short summary of what the robot is trying to do)

- Task Decomposition:  
Stage 1: (Action description + purpose)  
Stage 2: ...  
...

- Stage Classification:  
Stage 1: Auxiliary Stage / Interaction Stage  
Stage 2: ...  
...

- Keypoints by Stage:

Stage 1 Keypoints:  
Selected Keypoints: (e.g., 2)  
Operation: (e.g., grasp keypoint 2)  
Constraints: None

Stage 2 Keypoints:  
Selected Keypoints: (e.g., 4, 7)  
Operation: insert rigid part at 4 into soft object at 7  
Constraints: (4,7)

Final Reminders

- USE ONLY the colored dots and their associated visible numeric labels.
- If a keypoint is not marked this way, it does not exist.
- Maintain perfect consistency between visual grounding and reasoning.
\end{lstlisting}
% \end{tcolorbox}

% \subsection{Implementation Details of Point Tracker for Rigid Body}

% \begin{figure}[h]
%   \begin{center}
%    \includegraphics[width=1.0\linewidth]{img/prompt.pdf}
%   \end{center}
%   \vspace{-3mm}\caption{
%   \textbf{Error breakdown of
% the system modules.} 
%   }
%   \vspace{-3mm}
% \label{fig:prompt}
% \end{figure}

\subsection{Systems Error Breakdown}
The modular design of the framework entails an advantage for analyzing system errors due to its interpretability.
In this section, we perform an empirical investigation by manually inspecting the failure cases of the experiments reported in Tab.~\ref{tab:baseline}, which is then used to calculate the likelihood of a module causing an error while accounting for their temporal dependencies in the pipeline.
As reported in Fig.~\ref{fig:error}, the errors can be categorized into six main types:
\begin{itemize}
    \item VLMs: failures caused by incorrect task decomposition or keypoint selection;
    \item Sparse Correspondences: inaccurate sparse keypoint generation or mapping;
    \item Dense Correspondences: inaccurate tracking of keypoints on deformable objects;
    \item Point Tracker: inaccurate tracking of keypoints on rigid objects;
    \item Constraint Optimization: failures during the optimization process;
    \item Others: other issues, such as reachability limitations of the Franka arm
\end{itemize}

Among all modules, the 3D point tracker for rigid objects accounts for the largest portion of errors. This is primarily due to severe occlusions during rigid-deformable interactions, where thin rigid objects (e.g., hangers) are especially difficult to track reliably.
The second most significant source of errors comes from VLMs, which often fail to properly decompose long-horizon rigid-deformable interaction tasks or to select correct sparse keypoint candidates within each stage.
Errors from sparse and dense correspondences occur less frequently, but when they do, they are difficult to recover from due to the critical role of accurate correspondences in downstream processing.
Other modules, such as constraint optimization, contribute to some failures, but their impact is relatively minor compared to the core modules above.

% \subsection{Comprehensive Limitation Analysis}

% \paragraph{}

% \paragraph{}
% \subsection{Inference Results Example}

% Our work enhances robotic manipulation in tasks involving interactions between rigid and deformable objects by introducing a unified hybrid correspondence-based representation that combines sparse and dense keypoints. This approach offers significant societal benefits. It can empower assistive robots to better support the elderly and individuals with disabilities in daily living tasks, promoting independence and well-being. In household settings, it enables more practical and generalizable automation of complex tasks like hanging clothes and packing the racket. Additionally, the method has potential applications in industries such as textile manufacturing and food handling, where rigid-deformable interactions are common, leading to improved efficiency and safety. By capturing rich, task-specific interaction information in a generalizable and data-efficient way, this work represents a step forward in making intelligent, physically capable robots more accessible and useful in real-world environments.